\documentclass[runningheads]{llncs}
\usepackage{adjustbox}
 
\usepackage{eccv}

\usepackage{eccvabbrv}

\usepackage{graphicx}
\usepackage{booktabs}
\usepackage{subfiles}
\usepackage{comment}
\usepackage{wrapfig,lipsum}
\usepackage{multirow}
\usepackage[ruled,vlined]{algorithm2e}
\usepackage[accsupp]{axessibility}  

\usepackage[pagebackref,breaklinks,colorlinks,citecolor=eccvblue]{hyperref}

\usepackage{orcidlink}

\begin{document}
\newcommand{\sys}{\mbox{\textsc{$\nabla \tau$}}\xspace}
\title{\sys: \textsf{Gradient}-based and \\ \textsf{T}ask-\textsf{A}gnostic machine \textsf{U}nlearning}



\titlerunning{Robust Machine Unlearning via Aligning Loss Distributions}

\author{Daniel Trippa \and
Cesare Campagnano \and
Maria Sofia Bucarelli \and \\
Gabriele Tolomei \and 
Fabrizio Silvestri}

\authorrunning{D.~Trippa et al.}

\institute{Sapienza University of Rome \\
\email{name.surname@uniroma1.it}}

\maketitle


\begin{abstract}
\textit{Machine Unlearning}, the process of selectively eliminating the influence of certain data examples used during a model's training, has gained significant attention as a means for practitioners to comply with recent data protection regulations. 
However, existing unlearning methods face critical drawbacks, including their prohibitively high cost, often associated with a large number of hyperparameters, and the limitation of forgetting only relatively small data portions.
This often makes retraining the model from scratch a quicker and more effective solution.\\
In this study, we introduce Gradient-based and \textbf{T}ask-\textbf{A}gnostic machine \textbf{U}nlearning (\sys), an optimization framework designed to remove the influence of a subset of training data efficiently. It applies adaptive gradient ascent to the data to be forgotten while using standard gradient descent for the remaining data.
\sys offers multiple benefits over existing approaches. It enables the unlearning of large sections of the training dataset (up to 30\%).
It is versatile, supporting various unlearning tasks (such as subset forgetting or class removal) and applicable across different domains (images, text, etc.).
Importantly, \sys requires no hyperparameter adjustments, making it a more appealing option than retraining the model from scratch.
We evaluate our framework's effectiveness using a set of well-established Membership Inference Attack metrics, demonstrating up to 10\% enhancements in performance compared to state-of-the-art methods without compromising the original model's accuracy.
\end{abstract}



\section{Introduction}
\label{sec:intro}
The field of machine learning has seen remarkable advancements in the past years. Current state-of-the-art systems sometimes achieve performance levels comparable to human beings and obtain excellent accuracy in downstream tasks \cite{lecun2015deep, silver2016mastering, brown2020language, Gilardi_2023}. On the downside, deep machine learning models' growing complexity and scale introduce concerns about user privacy, potential biases in prediction, and deliberate manipulation of samples, among other issues \cite{shokri2017membership,yeom2018privacy, backdoorsurvey, biassurvey}. 
Indeed, publicly available models often rely on user-provided data and require adherence to the latest GDPR regulations, colloquially referred to as ``right to be forgotten'', allowing users to request the removal of their data from trained models for \textit{privacy} reasons.
Additionally, models trained on large amounts of publicly available data may encounter challenges in filtering and human-checking, potentially leading to \textit{biases} if the data contains toxic or inappropriate content  \cite{biassurvey}. 
Consequently, there is a need to remove these biases once they are discovered. 
Meanwhile, threat actors have a rising presence who purposefully manipulate training data with malicious intent, causing arbitrary mispredictions when certain patterns are detected. This model behavior, known as a \textit{backdoor} \cite{backdoorsurvey}, requires immediate removal of the influence of the manipulated data. All these scenarios, and many others, have in common the need to remove the influence of some training samples from a model.

A straightforward solution consists of discarding the model and starting a retraining procedure from scratch, excluding the data intended for removal. Because of the complexity of these models, this is often impractical, as the training procedures require a significant amount of time and computational power. 

Machine Unlearning has emerged as a critical field aiming to address this challenge efficiently, driving the exploration of new methodologies to mitigate the influence of training samples without resorting to the costly solution of complete retraining.

Current works in the literature focus on different variations of this task, using different definitions based on their objective. It is difficult to compare all these approaches, as there is a lack of a precise definition and specific metrics. One method may be suitable for a certain instance of unlearning, such as ``removing biases'' \cite{inproceedings}, but not work for other cases like ``removing backdoors'' \cite{Liu_2023_ICCV}.

In this context, we propose a comprehensive reinterpretation of the classic definition of approximating retraining, aiming for a more adaptable approach across various use cases. Our primary focus lies on enhancing User Privacy through unlearning. We emphasize the importance of defending against privacy-leaking attacks on specific training set samples.

Moreover, we aim to obtain a method that is not sensitive to the choice of hyperparameters and does not require extensive experiments to find the best setup, as this time and computation resources could be otherwise used directly for retraining. To achieve this goal, developing a process that remains independent of any specific hyperparameter is fundamental.

Considering the aforementioned issues, we aim to answer the following research questions. 
\textbf{RQ1}: Is an adaptive gradient ascent step a good approach to effectively remove training data information from deep learning models to enhance user privacy?
\textbf{RQ2}: How does the method perform in removing random samples of the training set, and how does it perform with different sizes of the subset to be removed? \textbf{RQ3}: How does the method perform in other settings  and domains (i.e. removing an entire class, image classification, text classification)?

To answer these questions, we introduce \textsf{\textbf{G}radient}-based and \textsf{\textbf{T}}ask-\textsf{\textbf{A}}gnostic machine  \textsf{\textbf{U}}nlearning (\sys ), an optimization framework designed to remove the influence of training data subsets efficiently. \sys demonstrates effectiveness across diverse unlearning scenarios while preserving model performance integrity. 
 We conduct extensive evaluations of \sys across different datasets and domains, including performing unlearning on models trained for image classification tasks on CIFAR-10 and CIFAR-100 \cite{cifar10dataset}, and text classification tasks on GoEmotion \cite{goemotions}.
Finally, we investigate \sys's performance across various hyperparameter values and various sizes of the set to be forgotten, demonstrating \sys robustness.

To summarize, the main contributions of this work are the following:
\begin{itemize}
    \item We present the first method that adapts the gradient step to the quantity of information to be forgotten. The procedure introduced is both Model and Task-agnostic. We demonstrate the superiority of our approach compared to the state-of-the-art methods. Our method can recover the accuracy levels present before the unlearning procedure.
    
    \item We conduct extensive experiments to empirically prove the effectiveness of our method and compare it with other approaches over a heterogeneous set of setups, including different domains (Text, Image), unlearning tasks (Random subset removal, Class removal) and sizes of the forget set, up to 30\% of the original training set;
    \item We perform a comprehensive evaluation of our method for different values of the sole hyperparameter introduced, providing insights on its correct use and empirically proving its robustness to small variations. 
\end{itemize}
Upon acceptance, we will publicly release our framework and code to reproduce the experiments.

\section{Related Work}
Several studies dive into the concept of unlearning. Existing works often focus on a particular subtask (e.g. removing bias, user privacy) and currently a standardized definition is missing.  Nguyen et al. \cite{nguyen2022survey} present a comprehensive survey where different aspects and open questions of machine unlearning are addressed.

\textbf{Unlearning for User Privacy}.
In this setting, the objective is to remove the influence of some subset of samples to protect the data against privacy leaking attacks, such as Membership Inference.
The current state-of-the-art for this particular task lacks consistency in its problem definition, framework, and evaluation methodology. Graves \textit{et al.} \cite{graves2020amnesiac} conceptualize unlearning as resistance to data-leakage attacks and perform \textit{label swapping} to remove information.
A more recent work by Chundawat \textit{et al.} \cite{chundawat2023bad} defines unlearning solely as \textit{removing} the information of forget set and use a randomly initialized \textit{incompetent teacher} to approximate the nature of the \textit{removed information}. Their focus consists in removing subclasses, where the samples to remove are not randomly selected, but share the same semantics (e.g. all images of a car). Foster \textit{et al.} \cite{foster2023fast} present a \textit{retraining free} method that aims to suppress the most influential weights for the forget set, but it only works for small enough forget set sizes (around 200 samples in the case of random subset removal). 
Kurmanji \textit{et al.} \cite{kurmanji2023unbounded} introduce an unlearning method based on bad teaching that works also in scenarios different from user privacy (such as removing biases). They discuss the resistance to privacy attacks, specifically Membership Inference Attacks, but evaluate their method only on class and subclass removal.

In these works, experiments frequently focus on subclass or class removal \cite{chundawat2023bad}\cite{kurmanji2023unbounded}; this setting not only limits the model's usability in real-world use-cases, such as the need to remove a user's data, but also complicates the assessment of whether the model has truly forgotten training examples. Instead, in studies exploring random subset removal, researchers often use a forget set size of less than 1\% \cite{foster2023fast}. This scenario prevents using these methods in contexts requiring larger datasets to be removed. 

Another limitation of these approaches is that they often introduce new hyperparameters that must be carefully selected before starting the unlearning procedure. This requirement limits the practical application of these approaches, as instead of fine-tuning their hyperparameters, it may be more time-efficient to perform simple retraining on the retain set.
Furthermore, it's worth noting that most existing studies are limited to models for image classification.

Our method performs effectively in scenarios involving random subset removal, where the samples designated for forgetting do not necessarily have a common class and do not necessarily share any similarity.
However, we do not confine ourselves to this scenario; we perform experiments and obtain optimal results also for class removal.
To meet the forget set size limitation, we show the effectiveness of our method on forget sets comprising up to 30\% of the original dataset.
To ensure our method is applicable to other domains, we test it in both image and text classification tasks.
Finally, our approach demonstrates remarkable robustness with respect to the sole hyperparameter we introduce, which is crucial for its practical applicability in real-world scenarios. As a part of our contribution, we also propose empirical insights on how to best configure this hyperparameter, only based on the size of the unlearning set.

\textbf{Membership Inference Attack}. Membership Inference Attacks (MIAs) \cite{yeom2018privacy,shokri2017membership}
are privacy attacks on machine learning models where an attacker tries to determine whether a sample was used during training. In the context of Machine Unlearning, this attack is used as a metric to determine if the unlearning procedure can protect User Privacy for a given Forget Sample \cite{nguyen2022survey}. In literature, there are different types of Membership Inference Attacks based on what the Attacker can observe. In this work, we use the same MIAs used by Kurmanji \textit{et al.} \cite{kurmanji2023unbounded} and Foster \textit{et al.} \cite{foster2023fast};
in this setting, the attacker only observes model outputs. This is sometimes referred to as Blackbox setting instead of Whitebox setting, where the attacker can access all model parameters. The attacker model is constructed by training a Logistic Classifier to differentiate between the output distribution of the Forget set and the Test set. Additional details will be included in \cref{sec:MIA}.


\section{Problem Definition}
\label{sec:problem_definition}

Let $D$ be a dataset and $\mathcal{A}$ a randomized training procedure. The output of a training procedure, given a Dataset $D$, and a fixed architecture is a vector of all model's parameters $ \mathcal{A}(D) = w_o$.  Given the internal randomness of the training procedure, $w_o$ is a random variable. 
We denote by $f( \cdot,w) $ the function implemented by the model having parameters $w$. 
 the following, with a little abuse of notation, we will often refer to a model as its parameters. Let's now define the \textit{forget set} as a subset $D_f \subset D$ of samples that we want to \textit{remove the influence} from the model. The \textit{retain set} is the complementary of the forget set, $D_r= D \setminus D_f$.
Given $w_o=\mathcal{A}(D)$, $D_f$ and $D_r$, the goal of a deep machine unlearning procedure $U$ is to produce a new set of weights
$w_u = U(w_o,D_{f})$ such that the ‘unlearned model’ $w_u$ has ‘forgotten’ $D_f$ without hurting the performance of the original model. By ‘forgetting' we mean the ability of the unlearned model to be indistinguishable under a certain metric from the golden baseline, which is the retraining only on $D_r$. 
In \textit{perfect unlearning} \cite{nguyen2022survey, bourtoule2020machine,brophy2021machine, thudi2022unrolling}, we ask for the distribution of the models trained only on $\mathcal{D}_r$ to be equal to the distribution of the unlearned model. Measuring these distributions is not trivial, and often, in specific applications, we can ask for a weaker unlearning definition, as we are not interested in preserving the weights distribution of the retrained model but only some of its proprieties.
Consequently, requiring strict equivalence between the parameter distributions of untrained and retrained models may prove impractical. Instead, a more feasible and interesting approach involves aligning the distributions of some metric computed on the model's weights.
More formally, for a map $M$ that takes as input the model $w$ and a subset of samples $X$ in the input space $ \mathbb{X}$ we require: 
\begin{equation}
\label{eq:weaker_unlearning}
 P ( M(X,w_u) \in S) = P  (M(X,w_r) \in S ) \;\; \forall S \subset \mathbb{M}, \; \forall X \subset \mathbb{X}.   
\end{equation}
When the map $M(X,w)$ is the output of the network having weights $w$ on samples $X$,  i.e., $M(X,w) = f(X,w)$, Eq. \ref{eq:weaker_unlearning} is also referred to as \textit{weak unlearning } \cite{baumhauer2022machine, nguyen2022survey}. 
In general, $M$ can represent any chosen mapping function or metric, depending on the desired properties of the model to be preserved. 
For example, when performing Unlearning for Bias removal, we are only interested 
in removing the influence of biased data, aiming for a result similar to the retrained model; in this setting, we would use a metric useful in measuring the level of bias in the model.
In the case of Unlearning for User Privacy, we like to evaluate the \textit{privacy} of samples to be removed  $x \in D_f$ after the unlearning procedure and the \textit{privacy} of the same samples in the retrained model. Ideally, we want this metric to match. In our setting, we use the Membership Inference Attacks (MIA) metric as a privacy metric. A full explanation of this metric can be found in \cref{sec:MIA}



\section{Our Method}
\label{sec:method}

To answer \textbf{RQ1}, our research introduces a novel loss function aimed at eliminating the influence of samples in the forget set while maintaining the integrity of the model's performance. 
With a primary emphasis on user privacy, our goal is to align the output distribution of samples within the forget set with that of the Test Set. %
We assume access to a validation set; more precisely, we only require access to aggregate information of the validation set: its mean loss value.  Discussions regarding this assumption are addressed in Section \ref{sec:limitations}.
Let  $L_D $ be the mean of the losses on a set $D$: $L_D = \frac{1}{|D|}\sum_{x\in D} l(f(x,w_i))$. We represent the mean loss on the forget set, retain set, and the validation set as $L_{D_f}$, $L_{D_r}$, and $L_{D_v}$, respectively. 
The loss function we introduce in our method is: 
\begin{equation}
L= \alpha ( \text{ReLU}(L_{D_v} - L_{D_f}))^2 + (1-\alpha) L_{D_r}.
\end{equation}
The term $ \text{ReLU}(L_{D_v} - L_{D_f})$ is used to reverse the gradient step on the forget set. This occurs only if the loss on the forget set is smaller than the objective loss $L_{D_v}$. Otherwise, the ReLU activation ensures that this term and its gradient become null and do not affect the optimization further.
Optionally, the objective loss can be recomputed every $c$ epochs.
The parameter $\alpha$ balances the noise injection and the fine-tuning term, $\alpha $ equal to $0$ corresponds to simple fine-tuning. 
Our findings suggest optimizing $\alpha$ using a scheduler yields the best results. Specifically, linearly decreasing $\alpha$ based on the number of optimization steps proves to be both efficient and fairly independent of the initial value of $\alpha$.
In Section \ref{sec:method_robustness}, we experimentally demonstrate how to select an appropriate $\alpha$ based on the size of the forget set relative to the retain set. Deriving the loss we obtain:
\looseness -1
\begin{equation}
\nabla L =
\begin{cases}
  - \alpha 2(L_{D_v} - L_{D_f}) \nabla L_{D_f} + (1-\alpha) \nabla L_{D_r}
  & L_{D_f} \leq L_{D_v}\\
  (1-\alpha) L_{D_r}
  & L_{D_f} > L_{D_v}\\
\end{cases}
\end{equation}
The squared term $\text{ReLU}(L_{D_v} - L_{D_f})^2$ scales the negative gradient step $\nabla L_{D_f}$ in proportion to $L_{D_v} - L_{D_f}$, effectively creating an adaptive step.\\
\textbf{Implementation insights: Balancing Retain and forget set}
An outline of our framework can be found in \cref{alg:our_method}.
In practice, the forget set is typically smaller than the retain set. Each optimization step in our method accesses a batch from both sets, with the same batch size.  Consequently, due to their size disparity, the Forget epoch —representing the number of steps required to process all forget set batches — differs from an epoch on the retain set.

\begin{adjustbox}{width=0.9\textwidth,center}
\vspace{1mm}
\centering
\begin{algorithm}[H]
\SetAlgoLined
\KwIn{Model, forget set, Validation Set, retain set,$\alpha$}
\KwOut{Updated Model}

\BlankLine
\SetKwFunction{ComputeMeanLoss}{ComputeMeanLoss}
\SetKwFunction{OptimizeLoss}{OptimizeLoss}
\SetKwFunction{Loss}{Loss}
\SetKwFunction{NextRetainBatch}{NextRetainBatch}
\SetKwFunction{OptimizeStep}{OptimizationStep}
\SetKwFunction{ReLU}{ReLU}
\SetKwFunction{ScheulerStep}{SchedulerStep}

\BlankLine

\For{each forget epoch}{
    \If{n\_epoch \% c == 0}{ 
        $L_{D_v}$ = \ComputeMeanLoss{Validation Set};
    } 
    \For{each $X_f$,$Y_f$ in forget set}{
        $X_r$,$Y_r$ = \NextRetainBatch{};
        
        $L_{D_f}$ = \Loss{$X_f$,$Y_f$};
        
        $L_{D_r}$ = \Loss{$X_r$,$Y_r$};

        $L$ = $\alpha$\ReLU{$(L_{D_v}-L_{D_f})^2$} + $(1-\alpha)L_{D_r}$
        
        \OptimizeStep{Model,L};
        
    }

    $\alpha$ = \ScheulerStep{$\alpha$}
}

\KwRet{Model}\;

\caption{\sys Training Loop}
\label{alg:our_method}
\end{algorithm}

\end{adjustbox}

\section{Experimental Setup} 
Our main focus is the forgetting of a \textit{random} subset of training data for User Privacy. With random, we mean that the samples have no correlation with each other (e.g. samples having the same class). In a real-world Unlearning scenario, samples are often not chosen randomly but are, for instance, the aggregation of all the User data we want to remove. 

To validate our method across diverse settings and, thus, answer \textbf{RQ2}, we conduct experiments on three classification datasets sourced from different domains. To address \textbf{RQ3}, we perform a series of experiments to demonstrate the robustness of our method across different conditions.
The details of each experiment's configuration are described in detail in Sec. \ref{sec:experimental_results}, alongside each explored setting.
\subsection{Architectures}
To empirically prove that our method is model-agnostic, we use two different model architectures in our experiments. Specifically, we use ResNet18 \cite{resnet} for both the Image Classification tasks, following Chundawat \textit{et al.} \cite{chundawat2023bad} and Foster \textit{et al.} \cite{foster2023fast}. For Text Classification we use RoBERTa \cite{liu2019roberta} with a linear layer on top. ResNet18 was trained for at most 50 epochs with data augmentation while RoBERTa was trained only for 10 epochs. Additional details are included in the Appendix. Data augmentation results in a lower starting MIA score; indeed, it prevents overfitting and diminishes the differences between test and training output distributions, reducing vulnerability to the attacker model.
\subsection{Datasets}
We evaluate our method on different datasets over different domains. We employ the established CIFAR-10 and CIFAR-100 \cite{cifar10dataset} benchmarks for Image Classification, in line with Chundawat \textit{et al.} \cite{chundawat2023bad}. To test in the domain of Text Classification, we use GoEmotions \cite{goemotions}, a dataset that labels sentences over 26 different emotions.

\subsection{Baselines}
\label{sec:baselines}
Machine Unlearning settings vary across literature, depending on the objective (e.g. User Privacy, Bias Removal, etc.) and the characteristics of the samples to remove (e.g. Item Removal, Class Removal, etc.). We compare our method with current state-of-the-art methods that perform random subset removal, particularly \textit{SSD} \cite{foster2023fast} and \textit{Amnesiac} \cite{graves2020amnesiac}. When performing Class Removal, we compare with \textit{SCRUB} \cite{kurmanji2023unbounded} that was specifically tested in this setting. Additionally, we include the golden baseline of \textit{Retraining}, and simple \textit{Fine-tuning}. 
In Retraining, we train the model from scratch on the retain set only. In fine-tuning, we optimize a pre-trained model on the retain set. 
We refer to the starting model trained on the whole dataset as \textit{Original}.

\textit{Retraining} is performed for the same amount of epochs used to train the \textit{Original} model. 
Instead, to guarantee a fair comparison, \sys, \textit{Fine-tuning} and all the other baselines except SSD \cite{foster2023fast}(that is training free) are run for only a fraction of the steps of Retraning (1/10). This approach ensures the same number of model updates for our method and all the baselines.




\subsection{Metrics}

\subsubsection{MIAs}
\label{sec:MIA}
\begin{wrapfigure}{r}{0.5\textwidth}
     \centering
\includegraphics[width=0.5\textwidth]{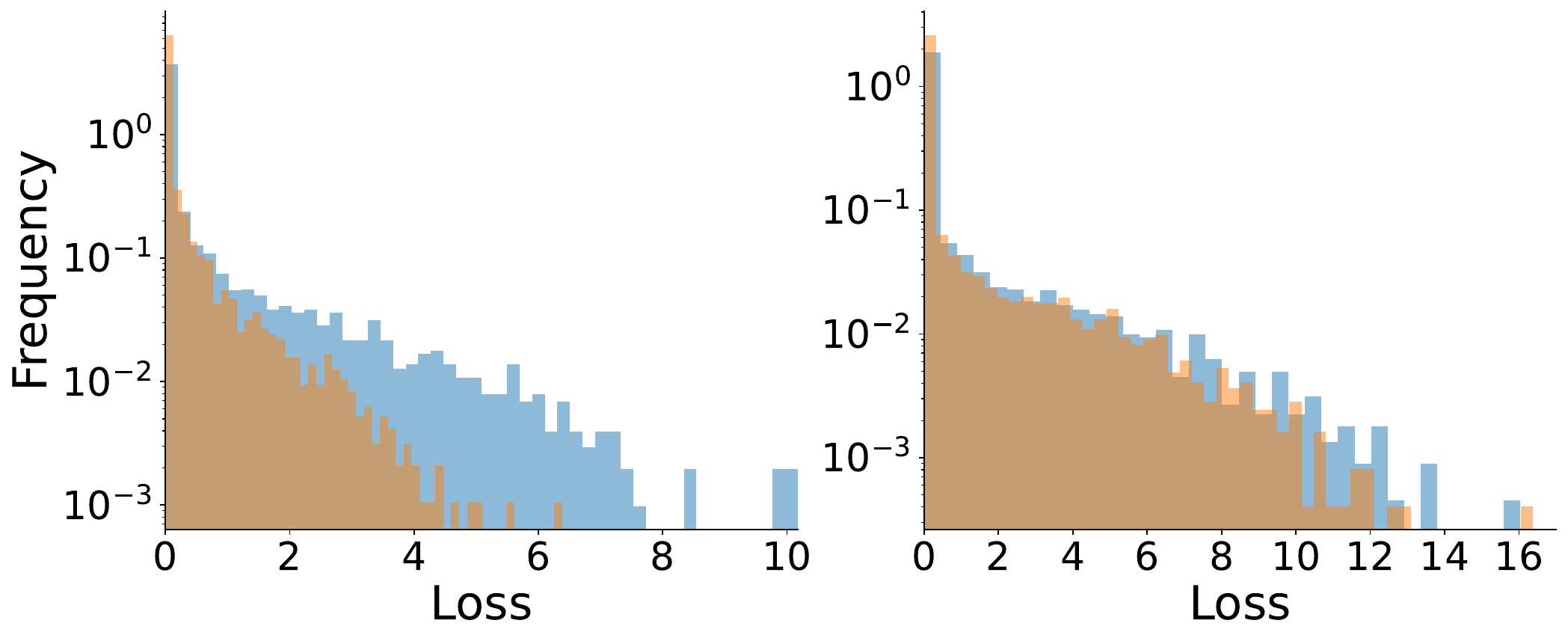}
     \caption{Loss distributions of Forget (in orange) and Test (in blue) sets before (left) and after (right) unlearning procedure on CIFAR-10.}
     \label{fig:lossdistpre}
 \end{wrapfigure}

To evaluate the privacy leaks of the forget set on the final model, we use Membership Inference Attacks (MIA) as a metric. 
In general, after training on the entire dataset, the distributions of losses and entropies of samples seen during training differ from those not seen during training (see left side of \cref{fig:lossdistpre}). This distinction allows the attacker to discern between the distributions and infer sample membership. We use the same MIA as Kurmanji \textit{et al.} \cite{kurmanji2023unbounded} and Foster \textit{et al.} \cite{foster2023fast}. Particularly, Kurmanji \textit{et al.} \cite{kurmanji2023unbounded} train the binary classifier using the unlearned model's losses on forget and test examples.
Conversely, Foster \textit{et al.} \cite{foster2023fast} utilize, as training data, the model's output entropies, defined as $E_s = -\sum_{i=1}^C p_{s,i}*log(p_{s,i})$, where $C$ represents the number of classes and $p_{s}$ denotes the network output for sample $s$ after softmax. In both cases, the objective is to classify forget versus test samples. 
We use the accuracy of an ‘attacker’ to evaluate machine unlearning. Specifically, we denote by $MIA_L$ the accuracy of an attacker trained using sample losses and by $MIA_E$ the accuracy of a model trained using output entropies. In the perfect scenario of Retraining the model only on $D_r$, the accuracy of the attacker model is equal to $50 \%$, which is equivalent to random guessing. In other words, the attacker cannot discern whether a sample belongs to the forget set or the Test Set (as, in fact, both were not seen during training).

\subsubsection{Accuracy}
Accuracy serves as a metric to assess both the efficiency and effectiveness of the final model. The Unlearning procedure must maintain high accuracy on the Test Set. It's worth noting that Accuracy also provides an indication of the model's ability to forget samples, typically demonstrated by lower accuracy on the forget set. Our goal is to achieve a forget set accuracy close to that of the Test Set.

\section{Experimental Results}
\label{sec:experimental_results}
We extensively test our method across various scenarios to evaluate its effectiveness. We focus on unlearning for user privacy, primarily in settings involving random subset removal, where the samples lack any specific correlation. Additionally, we conduct experiments in the case of class removal and empirically demonstrate the robustness of our method with respect to its hyperparameter. We will discuss each scenario separately and present the observed results.

\subsection{Defining optimal result}
The optimal result for machine unlearning varies depending on the specific setting.  In the case of User privacy,
the aim is to ensure that the samples designated for forgetting to maintain their privacy, making it impossible for an attacker to discern whether they belong to the test or forget set. Consequently, achieving an excessively low accuracy on the forget set and a MIA score of $0$ is not ideal. As highlighted by Golatkar \textit{et al.} \cite{golatkar2020eternal}, we could incur in the \textit{Streisand} effect: an excessively low accuracy could aid attackers in extracting information from the examined samples.
Hence, our goal is to maintain accuracy on $D_f$ close to that of the test set $D_t$, with the accuracy of the attacker model (that we called MIA metric) approaching that of random guessing ($50\%$).
For this reason, in the results we do not report the accuracy of the unlearned model on the forget set, but its absolute difference with the accuracy on the test set.

\subsection{Random Subset Removal}
In this Section, we aim at addressing \textbf{RQ2} by evaluating the performance of \sys in Random Subset Removal. In this setting, we randomly select some samples from the training set for removal. We test our method with different forget set sizes, namely 3\%, 15\%, and 30\% of the training set. 
We conducted a detailed comparison involving various methods: SSD \cite{foster2023fast}, SCRUB \cite{kurmanji2023unbounded}, Amnesiac \cite{graves2020amnesiac}, simple fine-tuning, and the retraining standard, both on CIFAR-10 and CIFAR-100. Results are shown in \cref{table:cifar10results} and  \cref{table:cifar100results}. It's important to note that, although both datasets involve the same task (Image Classification), there are significant differences between the two settings. Firstly, there is an order of magnitude difference in the number of classes. Additionally, CIFAR-10 exhibits a smaller initial difference between its test and forget sets, resulting in a lower starting MIA compared to CIFAR-100. 

As for \textbf{RQ3} we are also interested in different domains, we conduct the same experiments for a Text Classification task, as shown in \cref{table:nlpresults}. Due to the lack of state-of-the-art methods evaluated in this specific setting, we only compare our method against retraining and fine-tuning baselines.
\subsubsection{Results}
Our experiments revealed that for both Image Classification tasks, SSD (\cite{foster2023fast}) and Amnesiac \cite{graves2020amnesiac} did not effectively reduce the MIA accuracy on forget sets of the tested sizes. SCRUB \cite{kurmanji2023unbounded}, which focuses on class and subclass removal, demonstrates ineffectiveness in this setup. While Amnesiac \cite{graves2020amnesiac} effectively reduces the accuracy of the forget set, its impact on MIA accuracy is relatively minor compared to the original model, and in some cases, it is even worse than the original model (see experiment on CIFAR-10 with size of the forget set size of 3\%). 
In this setting, SSD \cite{foster2023fast} has no impact on accuracy and does not exhibit any noticeable effects on the MIA scores. 
As expected, fine-tuning leads to high accuracies on $D_r$ and $D_t$; however, this method does not give any warranty on forgetting and often shows bad performances on MIAs.
This outcome is anticipated since fine-tuning is executed on $D_r$, enabling the model to enhance its learning and performance specifically on that set. \\Comparatively, our method outperforms its counterparts in reducing the MIA score. In particular, \sys almost matches the retraining baselines for MIA scores both on loss and entropy distributions. Additionally, our approach maintains a consistently high Test Accuracy, even surpassing the initial accuracy in some cases, (such as in forget set sizes of 3\% and 15\% on CIFAR-10).\\
Text Classification results are collected in \cref{table:nlpresults}; we observe that our method is efficient in removing the influence of Forget Samples and defending against MIAs even if the starting model has a high difference between Train and Test set accuracy (27.3\%). It can be noticed that simple fine-tuning already defends well against MIAs in this particular setting.

\begin{table}[tb]
  \caption{Results on forgetting 3\%, 15\%, and 30\% of the CIFAR-10 train set. Mean and standard deviation values are averaged over three runs having different seeds. $A_D$ is the model's accuracy on dataset $D$. $MIA_L$ and $MIA_E$ denote the MIA score using loss and entropy distribution, respectively.  Bold font denotes the best results excluding the retraining golden baseline.  }
  \label{table:cifar10results}
  \centering
\resizebox{\textwidth}{!}{
\begin{tabular}{cl|c|c|c|c|c}
\hline
    &    &  $ A_{D_r} \uparrow$       & $   |A_{D_f} \hspace{-1mm}-\hspace{-1mm} A_{D_t} |\downarrow$       &  $ A_{D_t} \uparrow$       & |$MIA_L$ - 50| $\downarrow$       & |$MIA_E$ - 50 |  $\downarrow$       \\
\hline
 \multirow{7}{*}{\rotatebox[origin=c]{90}{forget set 3\%}} & Original   & 94.14± 0.00 & 9.24± 0.21  & 85.29± 0.21 & 5.13± 0.92  & 4.34± 1.02  \\
    & Fine-tuning & 99.75± 0.01 & 5.93± 0.67  & 85.67± 0.52 & 2.47± 0.85  & 2.03± 0.71  \\
       & Retraining  & 95.30± 0.02 & {0.70± 0.52}  & 84.07± 0.23 & {0.47± 0.31}  & {1.70± 0.87}  \\
    \cline{2-7}
    & SCRUB      & 94.11± 0.06 & 9.50± 0.10  & 85.19± 0.12 & 4.41± 0.60  & 3.50± 0.79  \\
    & SSD        & 94.14± 0.01 & 9.24± 0.20  & 85.31± 0.17 & 4.64± 0.64  & 3.44± 0.67  \\
    & Amnesiac   & 98.76± 0.09 & 18.87± 0.85 & 85.20± 0.16 & 11.20± 0.50 & 15.61± 0.51 \\
    & \sys (\textit{ours})       & \textbf{99.34± 0.07} & \textbf{2.06± 0.15}  & \textbf{85.94± 0.46} & \textbf{0.60± 0.59}  & \textbf{1.50± 0.70}  \\
    \hline \hline 
 \multirow{7}{*}{\rotatebox[origin=c]{90}{forget set 15\%}}& Original   & 94.14± 0.00 & 9.13± 0.21  & 85.29± 0.21 & 4.71± 0.28  & 3.76± 0.17  \\
    & Fine-tuning & \textbf{98.77± 0.02} & 6.30± 0.41  & 85.55± 0.30 & 3.02± 0.14  & \textbf{1.57± 0.33 } \\
    & Retraining  & 86.98± 0.39 & 0.44± 0.17  & 80.25± 0.55 & 0.49± 0.25  & 0.74± 0.25  \\
   \cline{2-7}
    & SCRUB      & 93.90± 0.14 & 9.10± 0.13  & 85.02± 0.31 & 4.86± 0.20  & 3.67± 0.21  \\
    & SSD        & 94.13± 0.02 & 9.07± 0.17  & 85.34± 0.18 & 4.88± 0.20  & 3.76± 0.65  \\
    & Amnesiac   & 96.73± 0.06 & 4.22± 0.69  & 84.77± 0.24 & 4.73± 0.42  & 11.15± 0.49 \\
    & \sys (\textit{ours})       & 97.82± 0.07 & \textbf{2.39± 0.11}  & \textbf{85.73± 0.25} & \textbf{1.52± 0.22}  & \textbf{1.72± 0.28}  \\
    \hline \hline
 \multirow{7}{*}{\rotatebox[origin=c]{90}{forget set 30\%}}& Original   & 94.14± 0.00 & 8.93± 0.21  & 85.29± 0.21 & 5.05± 0.24  & 3.47± 0.38  \\
    & Fine-tuning & 97.73± 0.05 & 6.72± 0.17  & 85.75± 0.33 & 2.90± 0.28  & 1.30± 0.18  \\
        & Retraining  & 95.17± 0.09 & 0.88± 0.40  & 82.24± 0.20 & 0.33± 0.39  & 0.46± 0.25  \\
    \cline{2-7}
    & SCRUB      & 94.09± 0.02 & 8.98± 0.24  & 85.17± 0.24 & 4.64± 0.12  & 3.63± 0.53  \\
    & SSD       & 94.14± 0.00 & 8.93± 0.21  & 85.29± 0.21 & 5.03± 0.13  & 3.62± 0.47  \\
    & Amnesiac   & 95.16± 0.07 & \textbf{1.39± 0.09}  & 84.55± 0.11 & \textbf{1.55± 0.92}  & 6.91± 0.23  \\
    & \sys (\textit{ours})       & \textbf{95.53± 0.12} & 3.31± 0.14  & \textbf{84.69± 0.05} & \textbf{1.70± 0.43}  & \textbf{1.35± 0.04}  \\
\hline
\end{tabular}
}
\end{table}

\begin{table}[tb]
  \caption{
Results for forgetting 3\%, 15\%, and 30\% of the CIFAR-100 train set across all baselines. 
Mean and standard deviation are obtained by averaging on three runs having different seeds.
   Bold font denotes the best results  excluding the retraining golden baseline.  $A_D$ is the model's accuracy on dataset $D$. $MIA_L$ and $MIA_E$ denote the MIA score using loss and entropy distribution, respectively.
  }
  \label{table:cifar100results}
  \centering
  \resizebox{\textwidth}{!}{
\begin{tabular}{cl|c|c|c|c|c}
\hline
    &    &  $ A_{D_r} \uparrow$        & $|A_{D_f}-A_{D_t}| \downarrow$       &  $ A_{D_t} \uparrow$       & |$MIA_L$ - 50| $\downarrow$       & |$MIA_E$ - 50 |  $\downarrow$       \\
\hline
   \multirow{7}{*}{\rotatebox[origin=c]{90}{forget set 3\%}}  & Original   & 99.21± 0.00 & 39.13± 0.10 & 60.13± 0.10 & 22.63± 0.07 & 20.03± 0.20 \\
    & Fine-tuning & \textbf{99.96± 0.00} & 31.22± 0.70 & \textbf{60.09± 0.25} & 15.22± 0.86 & 9.23± 0.77  \\
    & Retraining  & 98.77± 0.06 & 2.84± 0.68  & 53.49± 0.21 & 0.63± 0.53  & 0.13± 0.12  \\
    \cline{2-7}
    & SCRUB      & 96.65± 0.04 & 38.12± 0.36 & 58.33± 0.32 & 19.61± 0.51 & 16.42± 0.11 \\
    & SSD        & 98.82± 0.08 & 38.93± 0.16 & 59.94± 0.23 & 22.44± 0.22 & 19.97± 0.22 \\
    & Amnesiac   & 99.89± 0.01 & 34.28± 1.45 & 57.95± 0.28 & 11.98± 0.15 & 12.84± 0.08 \\
    & \sys (\textit{ours})       & 99.91± 0.00 & \textbf{3.78± 0.66}  & \textbf{59.35± 0.15} & \textbf{0.88± 0.52}  & \textbf{7.04± 0.59}  \\
    \hline \hline
    \multirow{7}{*}{\rotatebox[origin=c]{90}{forget set 15\%}} & Original   & 99.21± 0.00 & 39.12± 0.10 & 60.13± 0.10 & 24.24± 0.39 & 21.66± 0.32 \\
    & Fine-tuning & \textbf{99.97± 0.00} & 31.83± 0.32 & 59.65± 0.25 & 16.08± 0.53 & 10.15± 0.14 \\
    & Retraining  & 77.84± 2.73 & 0.50± 0.33  & 50.60± 1.11 & 0.73± 0.32  & 0.28± 0.18  \\
    \cline{2-7}
    & SCRUB      & 96.05± 0.33 & 37.83± 0.28 & 58.13± 0.08 & 21.05± 0.69 & 17.47± 0.71 \\
    & SSD        & 99.14± 0.07 & 38.97± 0.02 & \textbf{60.20± 0.03} & 23.90± 0.26 & 21.44± 0.49 \\
    & Amnesiac   & 99.81± 0.03 & \textbf{3.43± 1.28}  & 52.07± 0.33 & 5.37± 0.38  & 10.48± 0.54 \\
    & \sys (\textit{ours})       & 99.74± 0.01 & \textbf{3.38± 1.13}  & 58.39± 0.32 & \textbf{0.95± 0.35}  & \textbf{4.95± 0.50}  \\
    \hline \hline
   \multirow{7}{*}{\rotatebox[origin=c]{90}{forget set 30\%}}  & Original   & 99.24± 0.00 & 39.02± 0.10 & 60.13± 0.10 & 24.07± 0.37 & 21.39± 0.37 \\
       & Fine-tuning & \textbf{99.97± 0.00} & 32.11± 0.42 & \textbf{59.63± 0.28} & 16.32± 0.23 & 10.54± 0.23 \\
    & Retraining  & 99.22± 0.01 & 0.79± 0.48  & 49.22± 0.16 & 0.58± 0.13  & 0.63± 0.38  \\
    \cline{2-7}
    & SCRUB      & 96.85± 0.17 & 38.07± 0.14 & 58.48± 0.27 & 21.36± 0.49 & 17.97± 0.45 \\
    & SSD        & 99.24± 0.00 & 39.02± 0.10 & 60.13± 0.10 & 24.28± 0.44 & 21.34± 0.34 \\
    & Amnesiac   & 99.50± 0.07 & 17.32± 0.81 & 47.51± 0.35 & 11.62± 0.73 & 4.03± 0.36  \\
    & \sys (\textit{ours})       & 98.02± 0.09 & \textbf{5.89± 0.16}  & 54.48± 0.17 & \textbf{2.74± 0.37}  & \textbf{3.46± 0.18}  \\
\hline

\end{tabular}
}
\end{table}

\begin{table}[tb]
  \caption{Results on forgetting 3\%, 15\% and 30\% of train set on GoEmotion for Text Classification. We compare with the golden baseline of retraining and simple fine-tuning.
  Results are obtained by averaging across three runs having different seeds.   
   Bold font denotes the best results  excluding the retraining golden baseline. 
   $A_D$ is the model's accuracy on dataset $D$. $MIA_L$ and $MIA_E$ denote the MIA score using loss and entropy distribution, respectively.
  }
  \label{table:nlpresults}
  \centering
   \resizebox{\textwidth}{!}{
\begin{tabular}{cl|c|c|c|c|c}
\hline
    &    &  $ A_{D_r} \uparrow$        & $|A_{D_f}-A_{D_t}| \downarrow$       &  $ A_{D_t} \uparrow$       & |$MIA_L$ - 50| $\downarrow$       & |$MIA_E$ - 50 |  $\downarrow$       \\
\hline
   \multirow{4}{*}{\rotatebox[origin=c]{90}{\shortstack{forget set \\3\%}}}  & Original  & 82.83± 0.03 & 29.00± 0.24 & 54.59± 0.08 & 15.13± 1.31 & 7.56± 1.73 \\
       & Fine-tuning  & \textbf{61.17± 1.14} & \textbf{2.17± 1.25}  & \textbf{53.87± 0.61} & \textbf{1.17± 0.93}  & 1.79± 0.94 \\  
    & Retraining & 85.82± 0.26 & 2.04± 0.92  & 54.45± 0.16 & 1.48± 0.83  & 1.24± 0.66 \\
     \cline{2-7}
    & \sys (\textit{ours})      & 57.36± 1.83 & 2.84± 1.46  & 50.92± 1.15 & 1.29± 0.97  & \textbf{1.19± 0.70} \\
    \hline \hline
   \multirow{4}{*}{\rotatebox[origin=c]{90}{\shortstack{forget set \\15\%}}}  & Original  & 82.85± 0.04 & 28.62± 0.38 & 54.61± 0.42 & 15.20± 0.65 & 7.19± 0.09 \\
     & Fine-tuning & \textbf{60.92± 0.43} & \textbf{3.72± 0.84}  & \textbf{53.64± 0.64} & \textbf{1.18± 0.33}  & 1.08± 0.10 \\
    & Retraining & 85.77± 0.12 & 0.63± 0.38  & 54.73± 0.12 & 0.39± 0.29  & 0.43± 0.18 \\
     \cline{2-7}
    & \sys (\textit{ours})      & \textbf{59.34± 0.83} & 4.24± 2.18  & 52.17± 0.61 & 1.83± 0.97  & \textbf{0.70± 0.17} \\
    \hline \hline
    \multirow{4}{*}{\rotatebox[origin=c]{90}{\shortstack{forget set \\30\%}}} & Original  & 82.95± 0.02 & 28.36± 0.16 & 54.44± 0.20 & 15.01± 0.55 & 7.32± 0.16 \\
     & Fine-tuning  & \textbf{62.85± 0.33} & \textbf{5.27± 0.23}  & 52.96± 0.22 & \textbf{2.68± 0.06}  & \textbf{0.33± 0.14} \\
    & Retraining & 84.80± 0.20 & 0.32± 0.34  & 54.17± 0.15 & 0.38± 0.35  & 0.44± 0.24 \\
     \cline{2-7}
    & \sys (\textit{ours})      & \textbf{62.59± 0.92} & 6.32± 1.07  & \textbf{53.22± 0.92} & 2.93± 0.14  & 0.90± 0.48 \\
\hline

\end{tabular}}
\end{table}

\subsection{Class Removal}
As a complementary discussion to answer \textbf{RQ3}, we test how our method performs in a different setting. Specifically, we conduct experiments focused on removing the influence of an entire class of CIFAR10. We compare against SCRUB \cite{kurmanji2023unbounded}, a state-of-the-art method specifically designed for this task. We test for two classes: ``automobile'' (Class 1) and  ``dog'' (Class 5). 
We exclude the forgotten class from the Test set to evaluate the accuracy.
The MIA is computed on the loss (and entropy) only on samples belonging to the class we intend to forget, including both samples seen and unseen during training.

\subsubsection{Results}
The results presented in \cref{table:classremresults} demonstrate our approach's capability to entirely eliminate a specific class' impact. Accuracy on the forgotten class drops to 0, reproducing the results achieved through fine-tuning and the retraining baseline. The test accuracy remains consistently high, even surpassing the initial accuracy for class 5. Remarkably, the MIA score of our method notably decreases with respect to the original model, and \sys outperforms SCRUB in almost all setups.

\begin{table}[tb]
  \caption{Results on forgetting an entire class (``automobile'' and ``dog'') of CIFAR10 for Text Classification. 
 Mean and standard deviation are obtained by averaging across three runs with 3 seeds.
  Bold font denotes the best results, excluding the retraining golden baseline. 
  $A_D$ is the model's accuracy on dataset $D$. $MIA_L$ and $MIA_E$ denote the MIA score using loss and entropy distribution, respectively.
  }
  \label{table:classremresults}
  \centering
\resizebox{\textwidth}{!}{
\begin{tabular}{cl|c|c|c|c|c}

\hline
    &    &  $ A_{D_r} \uparrow$        & $A_{D_f} \downarrow$       &  $ A_{D_t} \uparrow$       & |$MIA_L$ - 50| $\downarrow$       & $MIA_E$ - 50 |  $\downarrow$       \\
\hline
   \multirow{5}{*}{\rotatebox[origin=c]{90}{Class 1}} & Original  & 93.89± 0.00  & 96.46± 0.00 & 84.56± 0.22  & 2.75± 1.46 & 3.04± 0.98 \\
       & Fine-tuning & 100.00± 0.00 & 0.00± 0.00  & 85.02± 0.20  & 1.68± 1.31 & 2.40± 1.38 \\
    & Retraining & 100.00± 0.00 & 0.00± 0.00  & 76.46± 0.48  & 1.23± 1.01 & 1.39± 1.10 \\
     \cline{2-7}
    & SCRUB     & \textbf{93.58± 0.10}  & 95.31± 0.14 & \textbf{84.19± 0.22}  & 3.07± 0.30 & \textbf{1.10± 0.89} \\
    & \sys (\textit{ours})      & 86.13± 18.50 & \textbf{0.00± 0.00}  & 76.41± 13.18 & \textbf{1.49± 1.49} & 2.34± 0.67 \\ \hline \hline
   \multirow{5}{*}{\rotatebox[origin=c]{90}{Class 5}} & Original  & 94.90± 0.00  & 87.32± 0.00 & 86.49± 0.17  & 6.81± 0.47 & 2.55± 0.64 \\
       & Fine-tuning & 100.00± 0.00 & 0.00± 0.00  & 88.21± 0.08  & 1.17± 0.62 & 1.06± 0.73 \\
    & Retraining & 100.00± 0.00 & 0.00± 0.00  & 80.64± 0.30  & 1.99± 1.00 & 0.98± 0.47 \\
     \cline{2-7}
    & SCRUB     & 94.40± 0.04  & 89.46± 0.27 & 85.85± 0.10  & 7.73± 0.89 & 3.34± 1.95 \\
    & \sys (\textit{ours})      & \textbf{99.82± 0.01}  & \textbf{0.00± 0.00}  & \textbf{88.55± 0.13}  & \textbf{2.14± 0.29} & \textbf{1.12± 0.80} \\
\hline
\end{tabular}
}
\end{table}

\subsection{Method Robustness}
\label{sec:method_robustness}
To test the robustness of our method, we repeat the procedure over many different forget set Sizes and with different values of our only hyperparameter $\alpha$. We repeat each experiment with three different seeds and showcase the mean value.  We plot the absolute difference from the perfect value 50\% in a heatmap shown in Figure \ref{fig:heatmaps}. A score close to 0 is a good defense against Membership Inference Attacks and approximates well the retraining baseline. Ideally, we want to observe a good score independently of the chosen hyperparameter $\alpha$.


\subsubsection{Results}

\begin{figure}[!h]
	\centering
	\begin{subfigure}{0.5\textwidth}
		\includegraphics[width=\linewidth]{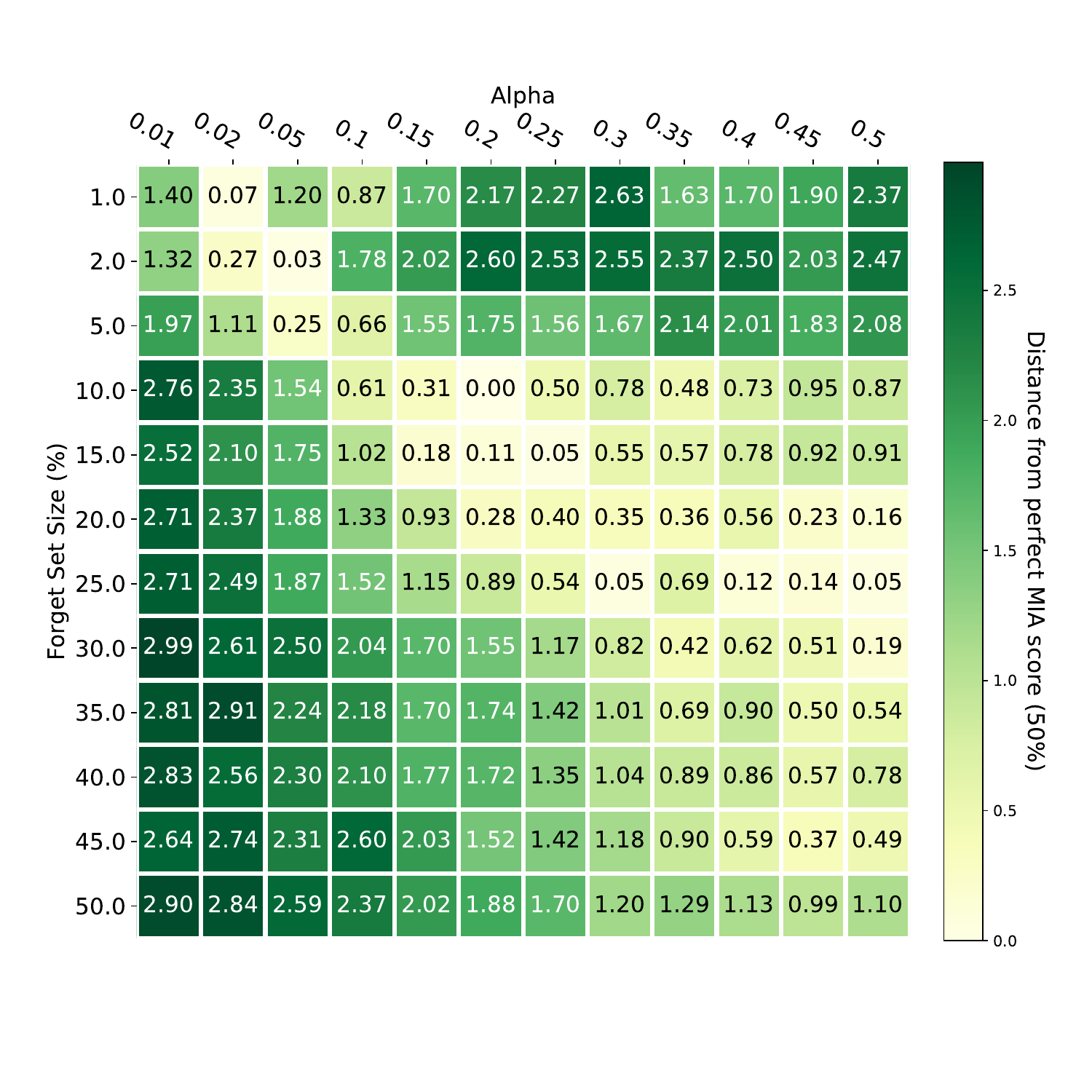}
		\caption{$|MIA_L -50| $}
		\label{fig:heatmapMIA}
	\end{subfigure}%
	\begin{subfigure}{0.5\textwidth}
		\includegraphics[width=\linewidth]{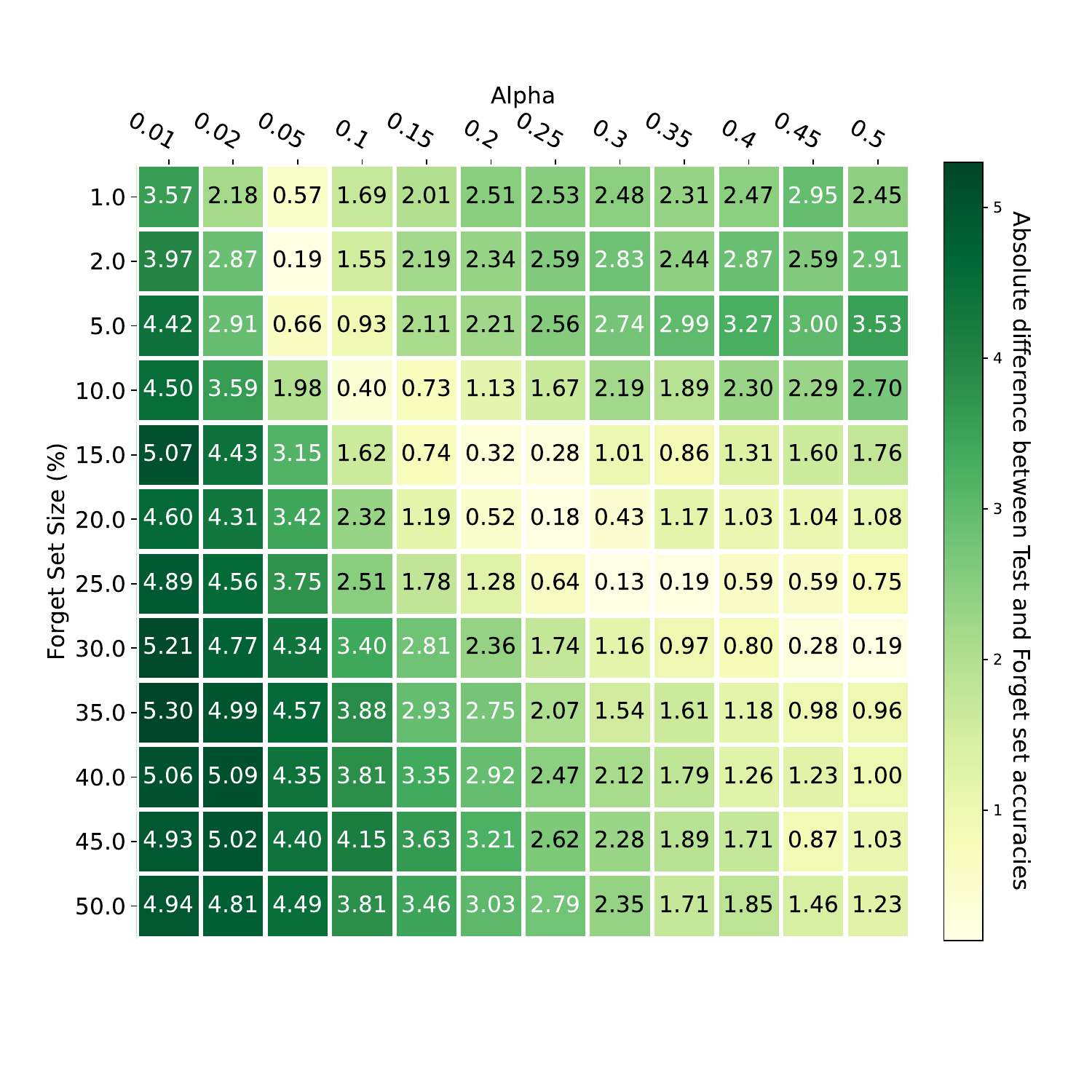}
		\caption{$|A_{D_f}-A_{D_t}|$ }
		\label{fig:heatmapACC}
	\end{subfigure}
   \caption{Experiment using our method on \textbf{CIFAR-10} across different forget set sizes (Y Axis) and $\alpha$ parameter (X Axis). On the left we report the absolute distance of $MIA_{L}$ from the ideal value 50\%. To improve the readability of the heatmaps, we do not report the standard deviations. However, for results where $|MIA_L - 50|< 1.0$ (the ones that best approximate retraining), the standard deviation is always under 1\%.  Note that even the golden baseline has a standard deviation  $\pm0.5\%$ from 50\%. On the right, we report the absolute difference between the accuracy on forget set $A_{D_f}$ and the accuracy on Test set $A_{D_t}$. The results highlight a similar pattern, indicating that similar scores in the accuracies are correlated to lower MIA scores. The results show the mean across three runs with different seeds.
  }
  \label{fig:heatmaps}
\end{figure}

As observed in \cref{fig:heatmapMIA}, for each split of $D_f$ there exist multiple values of $\alpha$ for which we obtain an optimal MIA score. This hyperparameter can be chosen in a range of values that all lead to good results. For instance, for a forget set size of $15\%$, all $\alpha$ values between $0.2$ and $0.45$ obtain values within the standard deviation of the golden baseline of retraining ($0.5)$. Even when choosing an $\alpha$ that is outside this range of values, it can be observed that the results are less than
$3\%$ away from the values obtained by retraining.

We also visualize the absolute difference between the accuracy on $D_f$ and the accuracy on $D_t$ in \cref{fig:heatmapACC} .
Consistent with MIA score observations, our method shows promise with sufficiently high $\alpha$. Examining these findings reveals a positive correlation between discrepancies in accuracy and MIA scores.

 As a rule of thumb, it can be derived by this experiment that a starting alpha that is around $\frac{5}{3}$ of the forget set size returns overall better performances across all the settings.


\section{Limitations}
\label{sec:limitations}
Our approach relies on having access to a validation set. 
However, even a tiny validation set, commonly accessible, suffices; indeed our method relies only on aggregate information of the validation set (average loss). 
If these are not available, one could treat the mean value of the validation losses as a hyperparameter that can be optimized, possibly starting from the average training loss.

 Second, despite being robust against the chosen $\alpha$, our method is still affected by classic hyperparameters, such as the used optimizer, the learning rate, and the scheduler. These hyperparameters can be estimated starting with the one used last in the training phase.

\section{Conclusions}
\label{sec:conclusions}
In this work, we introduced \sys, a novel approach for conducting Machine Unlearning, specifically targeting the removal of a substantial subset of training data influence from the final model. Our method places a primary emphasis on User Privacy, this is especially relevant because of ``the right to be forgotten'' of GDPR regulation, which requires data holders to be capable of removing user data upon request.

We demonstrate its optimal performance measured in terms of accuracy and MIA, where the latter serves as a more reliable way of ensuring the model has forgotten the desired data, compared to accuracy alone.
The effectiveness of our method has been validated across various settings, both in Image and Text classification tasks, outperforming several baselines. Notably, our approach showcases the capability to realign the distribution of the forget set closely with the Test Set, making our model excel in machine unlearning for User Privacy. Most importantly, the method we propose only introduces one hyperparameter and we show that the approach is fairly robust with respect to its starting value. This is crucial for an unlearning procedure, as to be applicable in real-world scenarios, it needs to be effective from the beginning of the procedure without requiring parameter tuning. 
In line with existing literature, we conducted experiments on unlearning models trained on classification tasks, including image and text classification.
However, our method is task-agnostic, relying solely on loss alignment, and could be applicable with slight modification across diverse downstream tasks. In future works, we aim to explore the adaptability of our method to additional tasks and domains.
Moving ahead, our hope is that this research will establish a new foundation for investigating unlearning with a focus on user privacy. Subsequent research should prioritize the standardization of Unlearning settings and definitions while exploring innovative methods adaptable to a variety of Unlearning scenarios.

\clearpage  

%
%

\bibliographystyle{splncs04}
\bibliography{main}

\clearpage
\appendix

\begin{center}
\end{center}

\section*{Hyperparameters}
\label{Appendix:structures_hyperparameters} 

\begin{table}[h]
\caption{Hyperparameters used during the training procedure on each Dataset. The resulting checkpoints are used for all following experiments on our unlearning procedure.}
\centering
\begin{tabular}{l|l|l|l}
\hline
              & \textbf{CIFAR-10}              & \textbf{CIFAR-100}              & \textbf{GoEmotions}          \\ \hline
\textbf{Architecture}  & ResNet18             & ResNet18              & RoBERTa             \\ \hline
\textbf{Optimizer}     & SGD                  & SGD                   & AdamW               \\ \hline
\textbf{Batch Size}    & 256                  & 256                   & 128                 \\ \hline
\textbf{Learning Rate} & 0.1                  & 0.1                   & 5e-5                \\ \hline
\textbf{LR decay}      & Linear {[}1,0.001{]} & Linear {[}1, 0.001{]} & Linear {[}1, 0.1{]} \\ \hline
\textbf{Weight Decay}  & 5e-4                 & 5e-4                  & 0.01                \\ \hline
\end{tabular}
\label{tab:hyperparameters}
\end{table}

\begin{table}[h]
\caption{Hyperparameters used during the unlearning procedure.}
\centering
\begin{tabular}{l|l|l|l|l}
\hline
              & \textbf{CIFAR-10}           & \textbf{CIFAR-100}          & \textbf{GoEmotions}        & \textbf{Class Removal}       \\ \hline
\textbf{Optimizer}     & AdamW              & AdamW              & AdamW             & AdamW               \\ \hline
\textbf{Batch Size}    & 256                & 256                & 128               & 256                 \\ \hline
\textbf{Learning Rate} & 0.001              & 0.001              & 0.0003            & 0.001               \\ \hline
\textbf{LR decay}      & Linear {[}1,0.1{]} & Linear {[}1,0.1{]} & N/A               & Linear {[}1, 0.1{]} \\ \hline
\textbf{Weight Decay}  & 0.01               & 0.01               & 0.01              & 0.01                \\ \hline
\textbf{Alpha}         & 5/3 * |$D_f$|      & 5/3 * |$D_f$|      & 0.5               & 5/3 * |$D_f$|       \\ \hline
\textbf{Alpha decay}   & Linear {[}1, 0{]}  & Linear {[}1, 0{]}  & Linear {[}1, 0{]} & Linear {[}1, 0{]}   \\ \hline
\end{tabular}
\label{tab:hyperparameters_exp}

\end{table}

To produce the starting checkpoint for the unlearning procedure on CIFAR-10 and CIFAR-100, we use data augmentations techniques including random crop and horizontal flip.
Interestingly, we noticed that using data augmentation techniques during training helps reducing the MIA score of the resulting model. We decided to keep these checkpoints, as they provide a more realistic setting for a real-world unlearning use-case.
All images are standardized by means and standard deviations. 

In all our unlearning experiments, we use the AdamW optimizer with weight decay. During pre-training and subsequent experiments, we always pick the last model produced by the optimization procedure.

 \cref{tab:hyperparameters} provides a  detailed description of the hyperparameters used in the pre-training procedures. The resulting checkpoints are used for all subsequent unlearning experiments.

In \cref{tab:hyperparameters_exp} we present the hyperparameters used to conduct the experiments on our unlearning procedure. Notice that the \textit{Retraining} baseline uses the same hyperparameters as pre-training, while \textit{Fine-tuning} employs the same hyperparameters as shown in \cref{tab:hyperparameters_exp} (except for alpha hyperparameter that is not present in the regular Cross-entropy loss). 

All the competing state-of-the-art baselines use the same hyperparameters provided by their official implementation. Our method and all the baselines, with exception of SSD, are based on a regular optimization procedure that runs unlearning for some epochs. SSD, instead, is a two steps \textit{retraining-free} approach. For a fair comparison, all the other baselines -- and our method -- run for the same number of steps, equivalent to 6 epochs on the retain set.

\end{document}